\documentclass[letterpaper, 10 pt, conference]{ieeeconf}  

\IEEEoverridecommandlockouts                              

\usepackage{hyperref}
\usepackage{siunitx}
\usepackage{graphicx}
\usepackage{float}
\usepackage{comment}
\usepackage[export]{adjustbox}

\def\eg{\emph{e.g.}} 
\def\etal{\emph{et al.}}
\usepackage{graphicx}
\usepackage{color,soul}
\usepackage{hyperref}
\usepackage{mwe}
\usepackage{subcaption}
\usepackage{amsmath}
\usepackage{tabularx} 
\usepackage{multirow}

\newcolumntype{L}{>{\centering\arraybackslash}m{3cm}}
\usepackage{float}
\usepackage[thinc]{esdiff}
\usepackage{graphics} 
\usepackage[table]{xcolor}
\usepackage{caption}
\usepackage{graphics} 
\usepackage{epsfig} 
\usepackage{mathptmx} 
\usepackage{times} 
\usepackage{amsmath} 
\usepackage{amssymb}  
\usepackage[noadjust]{cite}
\usepackage{booktabs}
\usepackage{multirow}
\usepackage{algorithmic}
\usepackage{ragged2e}
\usepackage{array}
\usepackage{textcomp}

\newcolumntype{P}[1]{>{\RaggedRight\arraybackslash}p{#1}}
\newcolumntype{M}[1]{>{\centering\arraybackslash}m{#1}}

\usepackage{microtype}
\renewcommand{\baselinestretch}{1.00}

\title{\LARGE \bf
Use-Inspired Mobile Robot to Improve Safety of \\ Building Retrofit Workforce in Constrained Spaces}

\author{Smruti Suresh, Michael Angelo Carvajal, Nathaniel Hanson, Ethan Holand, Samuel Hibbard, and Ta\c{s}k{\i}n Pad{\i}r$^{*}$
\thanks{This work was supported by the results of Phase I of the Envelope Retrofit Opportunities for Building Optimization Technologies (E-ROBOT) Prize, an American-Made Challenge funded by the U.S. Department of Energy.}
\thanks{$^{*}$Correspondence: {\tt\footnotesize t.padir@northeastern.edu}} 
\thanks{All authors are with the Institute for Experiential Robotics, Northeastern University, Boston, MA 02120 USA.}
\thanks{Nathaniel Hanson is now also with the Lincoln Laboratory, Massachusetts Institute of Technology, Lexington, MA 02421 USA.}
\thanks{Ethan Holand is now also with the Robotics Institute, Carnegie Mellon University, Pittsburgh, PA 15213 USA.}
\thanks{Ta\c{s}k{\i}n Pad{\i}r holds concurrent appointments as an Amazon Scholar and as a Professor of Electrical and Computer Engineering at Northeastern University. This paper describes work performed at Northeastern University and is not associated with Amazon.com, Inc. or its affiliates.}
}

\begin{document}

\maketitle
\thispagestyle{empty}
\pagestyle{empty}

\begin{abstract}

The inspection of confined critical infrastructure such as attics or crawlspaces is challenging for human operators due to insufficient task space, limited visibility, and the presence of hazardous materials. This paper introduces a prototype of PARIS (Precision Application Robot for Inaccessible Spaces): a use-inspired teleoperated mobile robot manipulator system that was conceived, developed, and tested for---and selected as a Phase I winner of---the U.S. Department of Energy's E-ROBOT Prize. To improve the thermal efficiency of buildings, the PARIS platform supports: 1) teleoperated mapping and navigation, enabling the human operator to explore compact spaces; 2) inspection and sensing, facilitating the identification and localization of under-insulated areas; and 3) air-sealing targeted gaps and cracks through which thermal energy is lost. The resulting versatile platform can also be tailored for targeted application of treatments and remediation in constrained spaces.

\end{abstract}

\section{Introduction}  \label{sec:intro}


Approximately $75\%$ of the world's greenhouse gas (GHG) emissions result from the cumulative energy sector~\cite{IEA2021Net0}. Roughly $75\%$ of the electricity consumed by the United States (U.S.), and over one third of its overall GHG emissions, are caused by the country's building sector alone~\cite{DOE2015QTR}. About $50\%$ of the total energy consumed by a typical U.S.-based residential building is spent heating or cooling its interior, and an estimated $30$--$50\%$ of that energy is lost due to air leakage~\cite{Chan2012Analysis}. Although modern thermal insulation materials to reduce energy loss have been in use since the mid-1980s~\cite{Kośny2022Short}, approximately half of the $140$ million residential and commercial buildings surveyed in the 2022 U.S. Census were built \textit{prior to} 1980~\cite{Census2022Year}. It is also estimated that $56$ million of those older thermally inefficient structures will still be standing in 2050~\cite{JLL2022Retrofitting}---the year by which the U.S. and other countries have pledged to achieve ``net zero'' carbon emissions as per the 2015 Paris Climate Agreement~\cite{UN2021Net0}.

To meet said emissions reduction target in time, the U.S. must begin retrofitting at least $3\%$ of its thermally inefficient buildings per year---a rate $3x$ the current global average~\cite{JLL2022Retrofitting}. The loss of thermal energy in residential buildings due to air leakage through a home's unconditioned attic can be significantly reduced by applying spray foam insulation, yielding a more thermally stable attic~\cite{Grant2018Comparison}. However, accessing and navigating attics and crawlspaces of residential buildings for retrofitting purposes is often challenging or even dangerous for humans~\cite{OAR2016DIY}. Risks in air-sealing include 1) the presence of asbestos, mold, and/or lead paint; 2) fire hazards such as recessed lighting and chimney/exhaust flues; and 3) workers falling through the ceiling~\cite{AirSealingReport}. According to the U.S. Bureau of Labor Statistics, 1,030 workers died between 2011 and 2018 from occupational injuries involving a confined space~\cite{OSHAStats}---$126$ of those workers died from inhaling a harmful substance resulting in asphyxiation. Additionally, several retrofitting contracting agencies refuse to work on manufactured home attic insulation since attics are greatly confined by the presence of low-slope roofs~\cite{AtticInsulationReport}. Given current sealing technologies and confined spaces, danger to humans is inherent to attic retrofit processes~\cite{SealingHazardsOSHA}.

    \begin{figure}[t]
        \centering
        \includegraphics[width=1.0\linewidth]{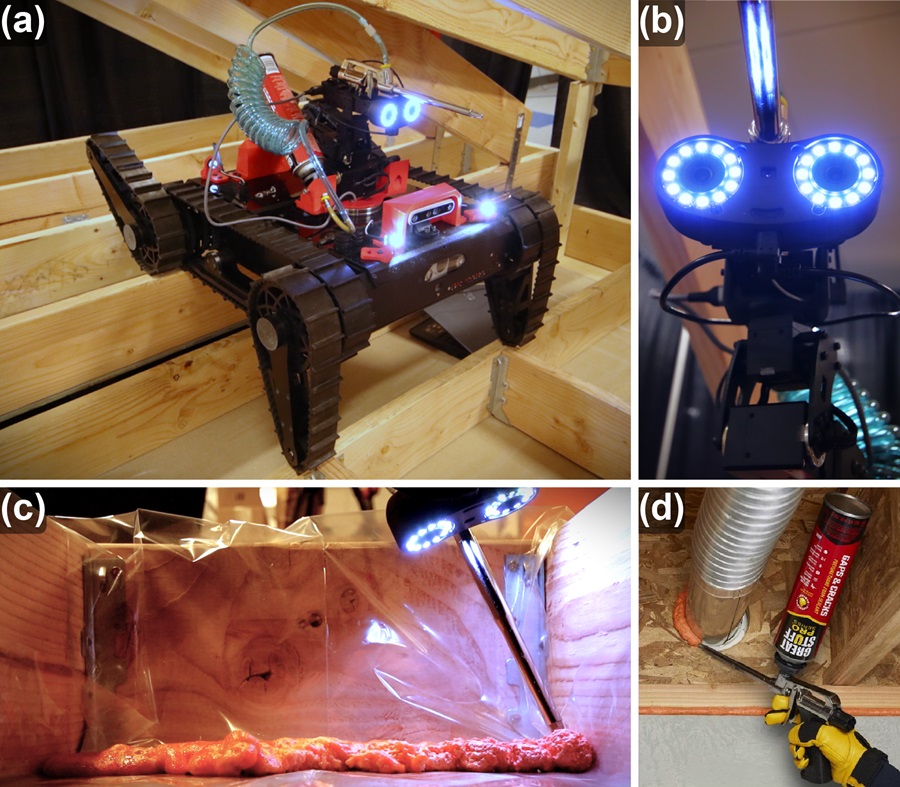}
        \caption{PARIS (a) traversing across attic testbed, (b) inspecting heat signatures, and (c) air-sealing a structural gap. (d) Human manually air-sealing fan duct with foam sealant (Great Stuff Pro\texttrademark)~\cite{DuPont2020GreatStuff}.}
        \label{fig:PARIS-Collage}
    \end{figure}

To help mitigate these risks and address the concerns of the workforce, the U.S. Department of Energy (DOE) convened the Envelope Retrofit Opportunities for Building Optimization Technologies (E-ROBOT) Prize, a nationwide $2$-phase challenge to stimulate the development of minimally invasive robotic solutions that make building retrofitting safer, easier, faster, and more accessible for human workers~\cite{DOE2020EROBOT}. This paper introduces \textbf{PARIS} (Precision Application Robot for Inaccessible Spaces): a mobile inspection and air-sealing robot for retrofitting thermally inefficient confined spaces,~\eg, residential attics, to increase levels of airtightness. A functional prototype of PARIS was fully built and tested during Phase II of the E-ROBOT Prize after DOE selected its concept and design as one of the ten monetary winners of Phase I.

PARIS' articulating mobile platform allows it to traverse the uneven terrain produced by the varying elevations of and spacing between attic floor joists. As it explores and creates a three-dimensional (3D) map of its operating environment, PARIS uses a temperature-sensitive camera to identify thermally lossy regions. PARIS air-seals the structural cracks and air gaps causing said thermal anomalies using a dispensing gun to precisely apply an insulating foam after first determining their location and geometry. PARIS contributes to the field of safety, security, and rescue robotics by:
    \begin{enumerate}
        \item Alleviating human workers from the risks of working in the hazardous confines of attics or crawlspaces;
        \item Analyzing readings from RGBD and thermal cameras that use sensor fusion to, respectively, generate 3D feature maps and identify gaps in need of air-sealing;
        \item Leveraging robotic control of spray foam sealant application for targeted sealing of thermally lossy gaps.
    \end{enumerate}

\section{Related Work}  \label{sec:related}

To motivate the design decision of PARIS, a variety of available robots and their suitability for operating within confined spaces were considered. Prior work with aerial robotic platforms in the context of building retrofitting applications focused on building inspection via external surveying~\cite{Xu2021Conceptual,Rakha2018Heat}. While highly maneuverable, airborne inspection robots may be limited by battery power, designated space for land and charge, and possible collision due to obstacle interference~\cite{Rakha2018Review}. Furthermore, maintaining flight stability while precisely controlling the intensity and direction of an interaction force, such as that generated by an air-sealing applicator, presents a significant challenge for aerial platforms~\cite{AerialRobotManipulation}. Therefore, airborne robotic solutions were not considered during PARIS' development.

A hexapod is a robotic mobility system that walks on six articulating legs and are known for their ability to traverse uneven terrain. Thus, Zang~\etal~\cite{Zang2023Perceptive} submitted an air-sealing solution built upon a commercial off-the-shelf (COTS) hexapod (SpiderPi, HiWonder)~\cite{HiWonderSpiderPi} to the E-ROBOT Prize and documented how said platform could also traverse across the joists of an attic floor. Once at the site requiring air-sealing, the sprayer-equipped hexapod dispenses foam sealant supplied to it through a hose connected to a remote containerized source. While this tethered setup allows the hexapod access to more sealant than it could carry independently, the hose’s fixed length ultimately restricts the robot's maneuverability and range within the attic.

Wheeled mobile robots have been developed for the purpose of autonomous air-sealing. Q-Bot is a four-wheeled remotely-controlled mobile robot that fits underneath the suspended floor of a residential home, sprays foam insulation to its underside~\cite{Lipinski2015QBot}, and autonomously decides where to spray foam next~\cite{Wagh2022Self}. As with~\cite{Zang2023Perceptive}'s solution, Q-Bot's maneuverability is limited because it too is tethered by a hose to an external supply tank that supplies insulation material. Additionally, wheeled robots would likely struggle to surmount the elevated joists commonly present in attics.

Tracked mobile platforms often traverse unknown field environments. For instance, one such robot with four articulating flippers (KIARA Telemax, AeroVironment) was employed to inspect radiation-exposed environments deemed to be dangerous for human personnel ~\cite{SuB2023Online}. The platform featured a multi-DOF manipulator arm that was retrofitted with a suite of cameras for navigating and mapping its environment, and a radiation sensor for identifying nearby radiation sources. The combination of its treaded mobility base, flippers, and articulated arm informed the design of the PARIS system.

\section{System Architecture}   \label{sec:architecture}

    \begin{figure}[b]
        \centering
        \includegraphics[width=1.0\linewidth]{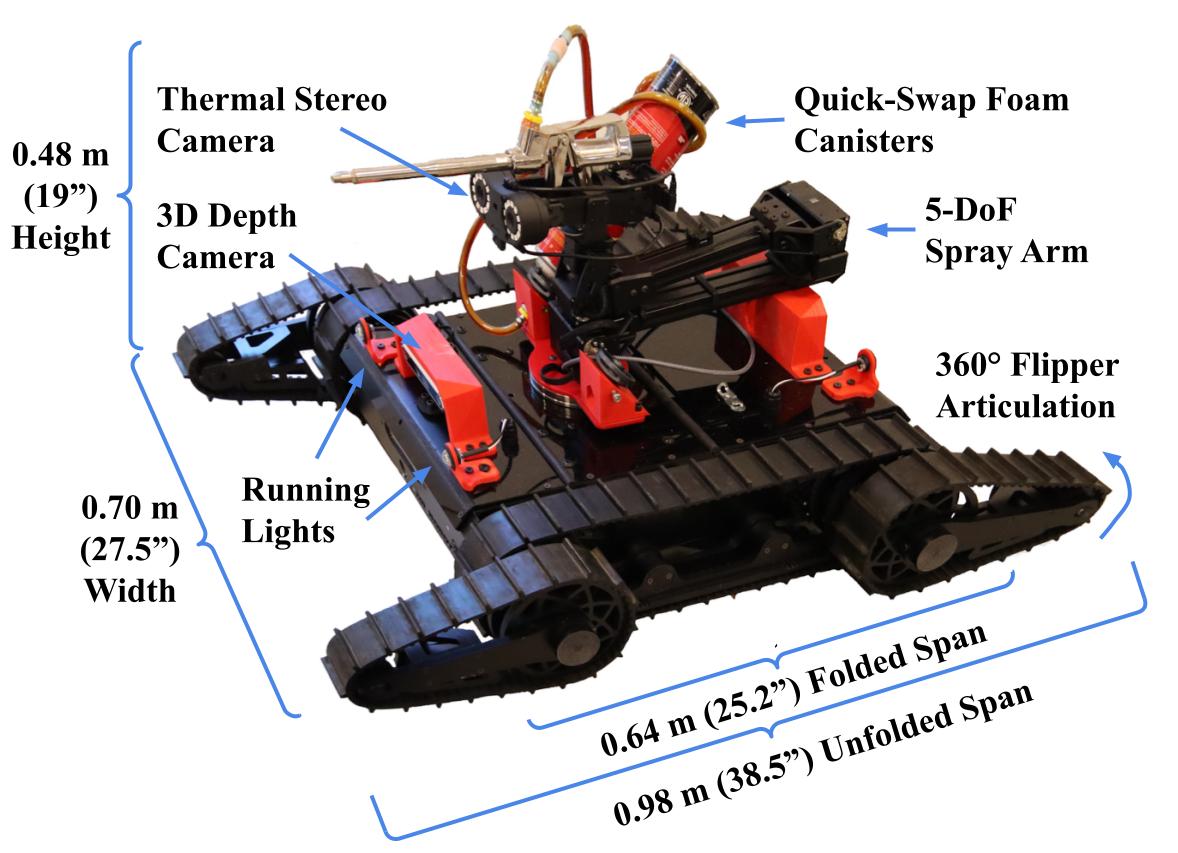}
        \caption{Physical prototype of PARIS as assembled for field testing.}
        \label{fig:paris1iso}
    \end{figure}

\subsection{Design Requirements}    \label{subsec:arch:reqs}
To benefit its human co-workers, PARIS must:
    \begin{enumerate}
        \item Fit through standard access hatches, which typically have a $57\times76$~\SI{}{\centi\meter} ($22.5\times30$ in) opening~\cite{ICC2024IRC};
        \item Traverse attic floors having elevation changes and structural joists spaced $30$--$60$~\SI{}{\centi\meter} ($12$--$24$ in) apart~\cite{ICC2024IRC};
        \item Generate a 3D map of its surroundings for reference;
        \item Discern the map's thermally lossy regions, if any;
        \item Apply foam air sealant to the targeted gaps and cracks.
    \end{enumerate}

\subsection{Prototype Hardware and Software}  \label{subsec:arch:specs}

\subsubsection{Locomotion}  \label{subsubsec:arch:specs:jaguar}
A COTS mobile robotic platform that consists of a center tracked section and four articulating flippers (Jaguar V4, Dr. Robot)~\cite{DrRobotJaguar} was chosen as the mobile base for PARIS' architecture because of its ability to angle its flippers to ascend or descend changes in the elevation of its surrounding terrain. Orienting the front flippers toward the rear ones, and vice versa, minimizes PARIS' overall length during transport; once placed down inside its operating environment, the flippers are rotated outward, as depicted in Fig.~\ref{fig:paris1iso}, to extend the robot's span; the larger span allows the robot to be supported by more weight-bearing joists during operation.

\subsubsection{Mapping} \label{subsubsec:arch:specs:map}
The PARIS base is equipped with a pair of depth cameras (RealSense 435i, Intel), one forward-facing and one rear-facing, each of which captures registered color (RGB) and depth (D) images in real-time. PARIS utilizes RTAB-Map~\cite{Labbé2018RTAB} to automatically create an online colorized 3D map of the operating environment by synchronizing and fusing the data from the parallel streaming video feeds.

\subsubsection{Manipulation}  \label{subsubsec:arch:specs:manipulate}
Mounted to the top of the PARIS base is a COTS $5$-DOF manipulator arm with a~\SI{750}{\gram} payload capacity (Interbotix ViperX 300, Trossen Robotics)~\cite{TrossenViperX}. To fit the~\SI{70}{\centi\meter}-wide PARIS through a standard attic access hatch, the integrated Interbotix arm is placed in its resting pose, reducing the robot's overall height to~\SI{48}{\centi\meter}.

\subsubsection{Inspection}  \label{subsubsec:arch:specs:inspect}
    
\begin{figure}[t]
        \vspace{0.5em}
        \centering
        \includegraphics[width=1\linewidth]{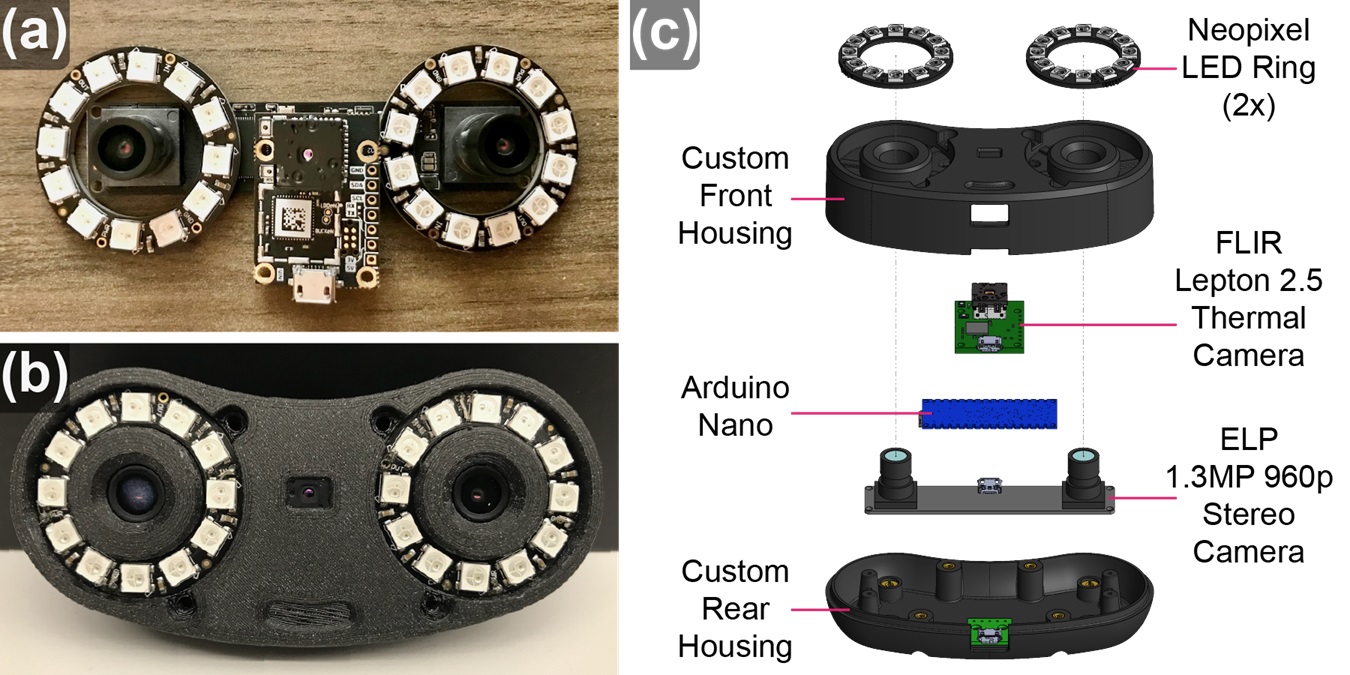}
        \caption{(a) Original conceptual component layout, (b) fully assembled state, and (c) exploded view of the thermal-stereo camera module.}
        \label{fig:spray_vision}
        \vspace{-0.5em}
    \end{figure}

The end-effector of PARIS' manipulator arm is also equipped with a custom composite thermal-stereo camera module, depicted in Fig.~\ref{fig:spray_vision}. The module integrates a stereo RGB camera (960P2CAM-V90, ELP), a pair of addressable LED light rings (NeoPixel Ring, Adafruit), and a centrally-located thermal imaging camera (Lepton, FLIR)~\cite{leptonFLIR}. Processing the pair of color images captured by the stereo camera component's RGB sensors yields a secondary 3D point cloud data of the scene. Due to the dexterity and reach of the 5-DOF arm supporting and maneuvering said composite module, the stereo camera can capture areas of the robot's taskspace beyond what PARIS' base-mounted RGBD cameras can see. By knowing the field of view (FOV), size, and known baseline distance between each camera sensor in its composite module, PARIS is able to map the thermal readings to the corresponding areas of said secondary point cloud. The resulting fused 3D thermal map facilitates the detection of areas in need of air-sealing by identifying steep temperature gradients and registers each as a Region of Interest (ROI) on the secondary map; example ROIs include a gap between a home's siding and its framing through which cold air enters an otherwise warm attic space. Later during the sealing process, the dual RGB sensors from the module's stereo camera offer PARIS' human operator a means of visual feedback, allowing the efficacy of the sealing attempt to be monitored.

\subsubsection{Air-Sealing} \label{subsubsec:arch:specs:sealing}
Canisters of Great Stuff Pro\texttrademark~(GSP)~\cite{DuPont2020GreatStuff}, a single-component polyurethane sprayed foam sealant, were mounted near the base of PARIS' manipulator arm through a quick-swap screw-on attachment. Each canister contains enough foam to cover a~\SI{1.3}{\square\meter} area. The pressurized foam compound is then piped from the canister's outlet to the inlet port of the GSP 14 Dispensing Gun~\cite{DuPont2020GreatStuff} retrofitted to the end of the arm. Early testing confirmed that the robot arm's advertised payload capacity could support the weight of the spray gun, its custom mounting bracket for retrofitting the gun to the arm, the custom camera module, the tube, and the insulating foam traveling through the tube. The last servo motor in the daisy chain of actuators along the length of the Interbotix arm, which originally controlled the arm's stock gripper end-effector, was repurposed to instead actuate the retrofit Dispensing Gun's trigger. When the mobile base is positioned near the target ROI, PARIS plans a waypoint trajectory for manipulating the dispensing gun as needed to apply the sealant to the geometry of the air leakage source. Once dispensed and fully cured, the water-resistant sealant becomes fire-retardant.

\subsubsection{Actuation and Power System}  \label{subsubsec:arch:specs:power}
The flippers on the mobile base are actuated by eight DC motors controlled via motor controller boards (SDC2130, Roboteq), which use encoders to enable both rotational speed and position control modes. The base is powered by \SI{20}{\ampere\hour} LiPo battery. DC buck converters were installed to create both a~\SI{19}{\volt} bus to power the Intel NUC and a~\SI{12}{\volt} bus to power the Interbotix arm and the mobile base's LED lighting; a~\SI{5}{\volt} USB hub powers the arm's custom thermal-stereo camera module.

\subsubsection{On-board Computing}  \label{subsubsec:arch:specs:compute}
\begin{figure*}[!bt]
        \centering
        \includegraphics[width=1.0\linewidth]{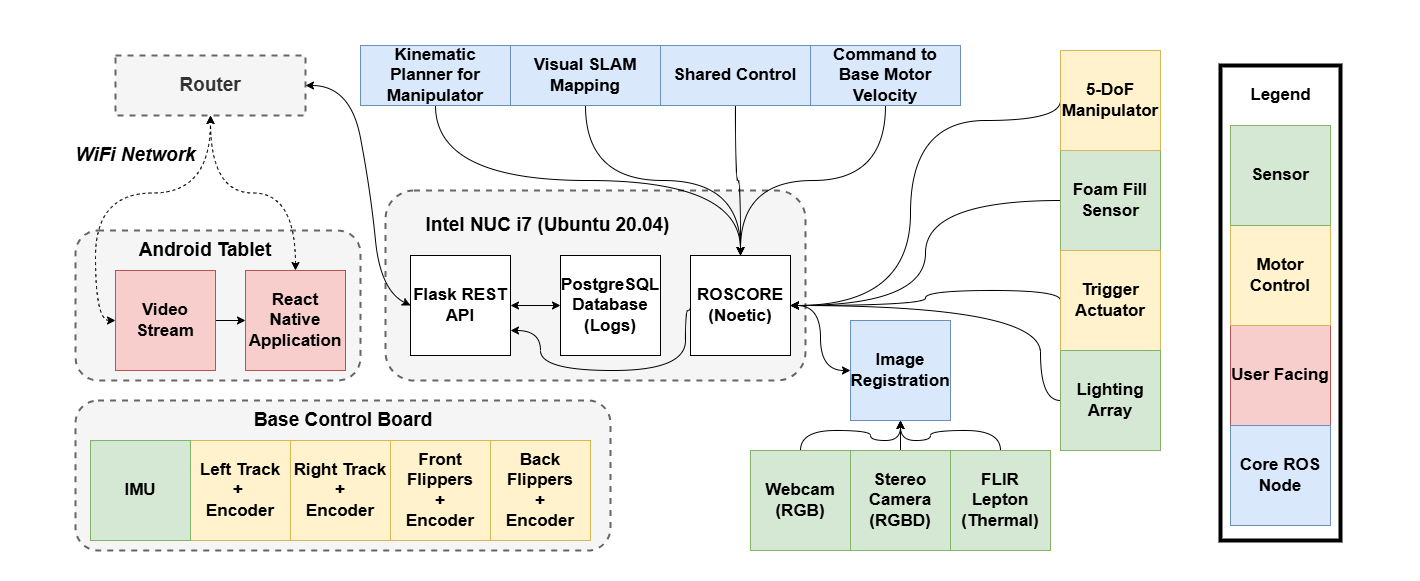}
        \caption{\textbf{Hardware-Software System Architecture.} PARIS operates using a ROS architecture system of nodes connecting over wired and wireless network connections to allow for sensor data transmission and user control.}
        \label{fig:paris-sys-arch}
        \vspace{-1em}
    \end{figure*}

PARIS uses an Intel NUC i7 computer kit as its primary processing unit; this NUC offers $4$--$6$ cores and $8$--$12$ threads of processing power; it also provides the flexibility to configure the system with more than~\SI{16}{\giga\byte} of RAM, if required. This facilitates efficient handling of complex robotic tasks such as motion-planning, and real-time point cloud generation. As illustrated in Fig.~\ref{fig:paris-sys-arch}, the Ubuntu 20.04-configured NUC runs the Robot Operating System (ROS)~\cite{quigley2009ROS} as a middleware to interconnect PARIS' sensors, algorithms, and actuators.

\subsubsection{Human Robot Interaction} \label{subsubsec:arch:specs:gui}
A cross-platform React Native mobile application was custom-designed for easy operation. As seen in Fig.~\ref{fig:teleop}, this touch-enabled application: 1) runs on a tablet device, 2) communicates in real-time with the robot, 3) displays key system status information,~\eg, signal strength and battery voltage, 4) visualizes data streams for the user, and 5) allows the user to teleoperate the robot via a virtual joystick and buttons rendered on the touchscreen. The user can choose between video feeds from the robot’s visual sensors and a visualization of the robot’s 3D map of the operating environment to monitor its progress in exploring the taskspace and supervise the air-sealing process for defects such as partially or inconsistently sealed air gaps. The app also allows the user to toggle between driving the robot around and controlling the arm and its spray gun. The app was built on a Flask server that connects to ROS, minimizing work to integrate new features with the application.

\section{System Evaluation}   \label{sec:evaluation}

\subsection{Test Environment}
\begin{figure}[b]
        \centering
        \includegraphics[width=1.0\linewidth]{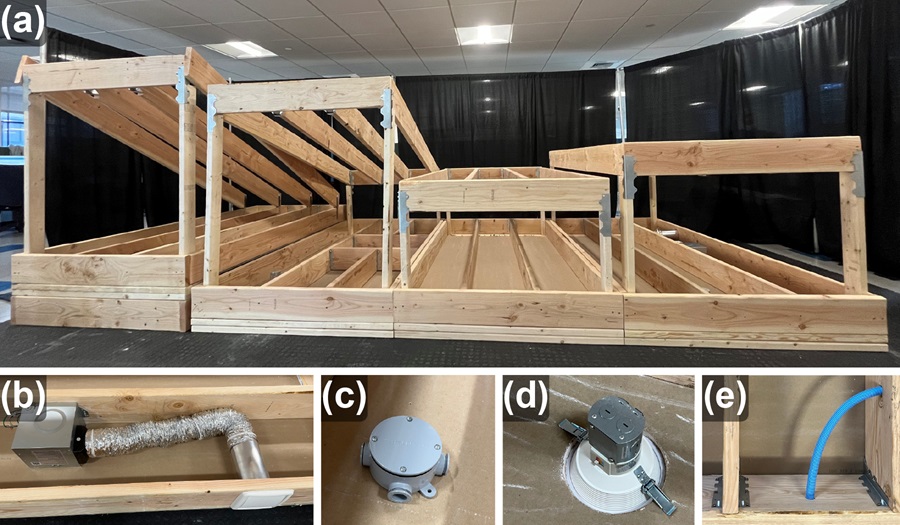}
        \caption{(a) Adjacent gable and flat roof testbed modules. (b) Fan duct, (c) junction box, (d) light fixture, and (e) electrical conduit.}
        \label{fig:attic-testbed}
    \end{figure}

To demonstrate the functionalities and features of PARIS, four testbeds that replicated different portions of a typical household-sized attic were constructed, each with footprint measuring $122\times251$~\SI{}{\centi\meter} ($48\times99$ in). Fig.~\ref{fig:attic-testbed}(a) depicts the modules placed side by side. Each testbed comprised of: 1) a floor with $35.5$--$40.5$~\SI{}{\centi\meter} ($14$--$16$ in) joist spacing; 2) a ceiling architecture with unique vertical clearances and roof pitches; and 3) at least one distinct structural feature in need of air-sealing~\eg, can lighting or bathroom fan. One testbed's floor also contained a standard size access hatch.


At the time of testing, each testbed approximated a simplified model of a typical section of an attic and was therefore free of common obstacles such as batt insulation, dirt, or wiring cables. However, it included many physical features of attic interiors that contain air gaps such as chimneys, recessed lights, bathroom fans, junction boxes, pipes, and ducts; the specific attic features tested during the development of PARIS are shown in Fig.~\ref{fig:attic-testbed}(b)--(e).

\subsection{Test Plan}
The plan for testing PARIS consisted of the following tasks: 1) sealing wall tops near the gable ends; 2) performing air-sealing in the low-flat roof (utilizing the space between rafters to extend the arm); 3) sealing the holes perpendicular to outer walls (\eg, gaps around conduit or wall vents); and 4) traversing across joists of varying elevations.

\subsection{Test Results}
\begin{figure}[t]
        \vspace{0.5em}
        \centering
        \includegraphics[width=1.0\linewidth]{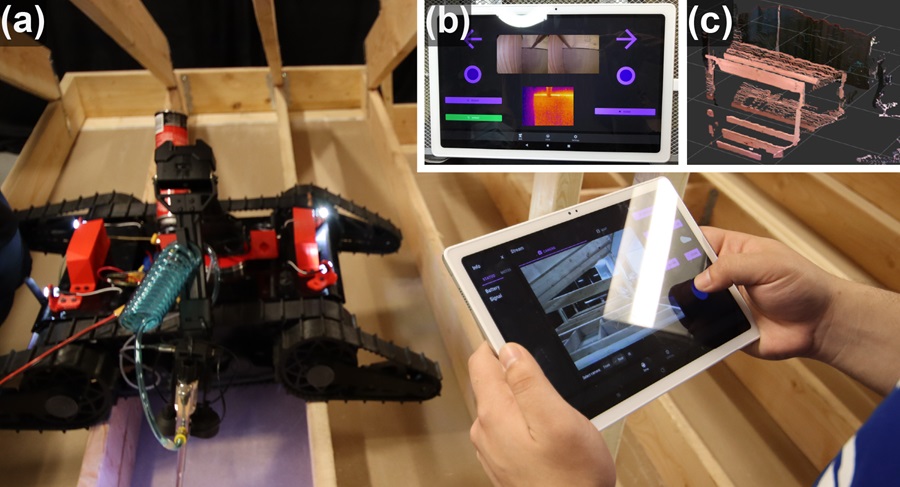}
        \caption{(a) Human operator controls PARIS via tablet device whose custom app displays visual information on demand. (b) Video streams from thermal-stereo module. (c) Internally stored 3D point cloud.}
        \label{fig:teleop}
    \end{figure}

\begin{figure}[t]
        \centering
        \includegraphics[width=1.0\linewidth]{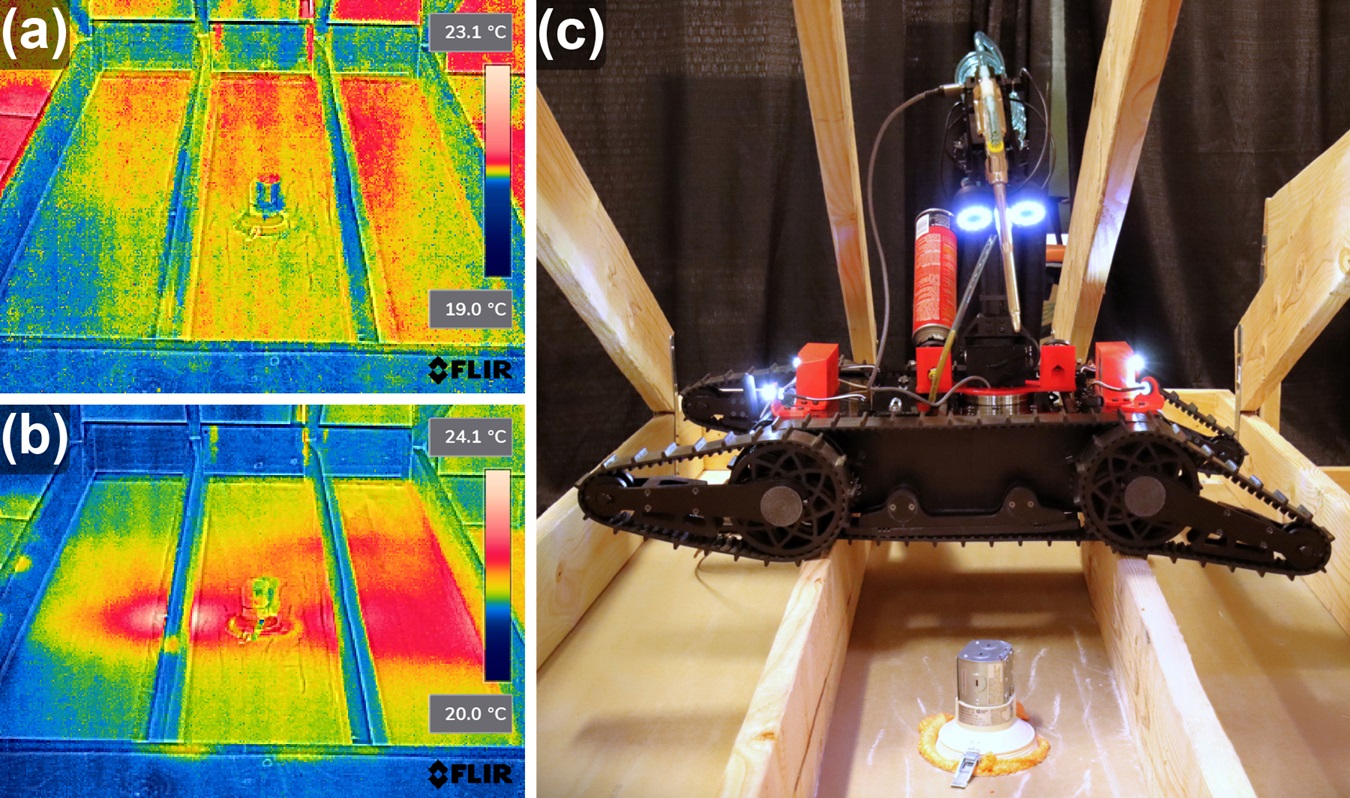}
        \caption{Thermal images of the attic test bed and light fixture (a) before and (b) after sealing. (c) PARIS beside the air-sealed fixture.}
        \label{fig:air-sealing}
        \vspace{-0.5em}
    \end{figure}

    \begin{enumerate}
        \item 
            \textbf{Mobility:} PARIS can traverse across joists spaced $35.5$--$40.5$~\SI{}{\centi\meter} ($14$--$16$ in) apart and surmount at least an~\SI{18}{\centi\meter} ($7$ in) step by articulating its flippers and maintaining contact with at least two joists at all times.
        \item 
            \textbf{Mapping:} PARIS generates a colorized 3D point cloud of its environment, as shown in Fig.~\ref{fig:teleop}(c), by utilizing RGBD cameras through RTAB-Map~\cite{Labbé2018RTAB}.
        \item 
            \textbf{Inspection:} PARIS' thermal-stereo camera module can capture RGB and thermal images of ROI in need of sealing, as seen in Fig.~\ref{fig:teleop}(b), which shows the video streams captured by said module as it stares at a structural gap with a thermal anomaly.
        \item 
            \textbf{Air-Sealing:} PARIS can successfully manipulate the Dispensing Gun to seal a targeted geometry such as the circumference of the light fixture that was installed in the floor of the attic testbed [see Fig.~\ref{fig:air-sealing}(c)]; post-application, the temperature of the floor around the fixture increased over a period of $30$ minutes because the foam sealant prevented the loss of thermal energy that was initially escaping; thermal photos taken by an external thermal camera (T420bx, FLIR) $15$ minutes before and after the GSP foam application are shown in Fig.~\ref{fig:air-sealing}(a) and Fig.~\ref{fig:air-sealing}(b), respectively.
        \item 
            \textbf{Human-Robot Collaboration:} Users control PARIS as shown in Fig.~\ref{fig:teleop}(a) and Fig.~\ref{fig:teleop}(b)  using the tablet-based user interface and monitor the robot's operation through sensor feeds from the thermal and RGBD cameras.
    \end{enumerate} 

\section{Discussion}    \label{sec:discussion}  

This section details some of the challenges and technical innovations that emerged during the design and development of this robotic platform.

\subsection{Robot Platform} \label{subsec:platform_discuss} 

Accurate odometry information for pose estimation was hard to obtain due to the motor encoders on the off-the-shelf mobile base. A TCP link to the proprietary board was successfully established, and encoder positions were occasionally polled, but the velocity information was not available. To diagnose the issue, the mobile base was fixed on a high platform such that the tracks did not touch the ground, and the motors could be continuously run for data collection without the mobile base colliding with its surroundings. Additionally, a simple MATLAB script to obtain raw encoder velocity values was utilized. The test showed that the raw encoder velocity values remained at zero despite the motors running at varying speeds. To address this, visual odometry was implemented using the RTAB Odometry node, which estimates odometry from RGB-D camera feeds. However, the drift inherent in visual odometry led to velocity overshoots, causing the robot to exceed its target position. To proceed with testing safely, AR tags were mounted on the testbed, and RTAB-Map parameters were adjusted to increase reliance on AR tag detection for loop closures and pose corrections. This solution was intended as a temporary fix, with plans to implement sensor fusion for accurate and stable pose estimation in the future.

\subsection{Sealing Methodology}    \label{subsec:sealing_discuss}

The Interbotix ViperX 300's original URDF (Unified Robot Description Format) text file for representing the robot arm's model was modified to include the position and orientation of the GSP 14 Dispensing Gun by referencing a DIY 3D model of the gun made by manually measuring the geometry of its critical features,~\eg, the distance between the gun's dispensing tip and its point of attachment to the Interbotix arm. Moveit~\cite{Chitta2016MoveIt} was used to control the arm and generate optimal waypoints for Cartesian Planning. Straight line trajectory planning was unreliable due to jitters caused by the motors in the arm when it was extended. Instead, semi-circle trajectory planning was chosen since three waypoints could be specified and the velocity to each waypoint could be controlled enabling a much more smoother movement. Motion planning was tested in simulation with Moveit and Rviz~\cite{Kam2015RViz} before testing it on the actual hardware. 

\section{Conclusion and Future Work}    \label{sec:conclusion}
The PARIS system, built with a COTS mobile base and robotic arm, demonstrated technical innovations that are core to a robot solution for constrained spaces and building retrofit tasks like air-sealing. On a full-scale attic testbed, the robot demonstrated that it can navigate the precarious environment on top of attic joists without falling using RTAB-Map SLAM powered by stereo cameras. The thermal camera on the end effector was shown to be able to identify areas to be air-sealed on the 3D map and the robotic arm can apply spray foam along straight and curved cracks and edges. Our purpose-built, cross-platform tablet interface provides a simple interface for maneuvering the robot and apply insulation, abstracting away the complexity of the robot’s 10+ degrees of freedom.

    \begin{table}[b]
        \centering
        \begin{tabular}{|c|c|c|} \hline
            \textbf{AFFECTED TRAIT}  &   \textbf{PARIS 1.0}  &   \textbf{PARIS 2.0}\\ \hline
            \textbf{Width [cm]}    &    $70.1$  & $55.9$\\
            \textbf{Span (Unfolded) [cm]}      &   $98$   &   $112.5$\\
            \textbf{Total Mass [kg]}   &   $30$    &   $21.8$\\
            \textbf{Cost of Goods [USD]}     &   $21,500$ &   $4,500$\\ \hline
        \end{tabular}
        \caption{Technical improvements across PARIS platforms}
        \label{tab:paris1vs2}
    \end{table}
    
Future work on PARIS involves migrating its sensors and actuators over to a custom-designed mobile base designed and fabricated by the Institute for Experiential Robotics at Northeastern University to better meet the system's product requirements. The architecture of the new mobile base increases the robot's span across neighboring joist when its flippers are rotated to extend outward, increasing PARIS' stability as it traverses across or parks itself within the attic space; the new design also reduces the overall width of the system from~\SI{70.1}{\centi\meter} ($27.6$ in) down to~\SI{55.9}{\centi\meter} ($22$ in), allowing it to be compact enough to pass through a standard attic hatch opening in more than $1$ orientation. As Table~\ref{tab:paris1vs2} outlines, this next-generation prototype is expected to both weigh and cost less to build than the original prototype. The new base will also feature quick-swappable flippers, allowing alternate drive configurations and reduced maintenance time.

The system will be made semi-autonomous with human-in-the-loop (HITL) where the operator solely inputs the locations that need to be air-sealed and PARIS will autonomously navigate to the location and perform air-sealing. The 5-DOF manipulator arm will be upgraded to a 6-DOF one which will allow PARIS to articulate the spray gun in more directions and at more angles as its nozzle traces the calculated trajectories planned by the system’s controller to best align with the gaps needing sealant. Appropriate state-of-the-art navigation and motion planning algorithms will be implemented to test PARIS' performance in a real attic that features challenging obstacles and operating conditions such as debris, dust, and moisture. The long-term goal for developing PARIS is to allow human workers to complete other building retrofitting tasks in parallel to it elsewhere, hence shortening a weatherization project’s time to completion.

\section*{ACKNOWLEDGMENTS}

The authors gratefully thank all Northeastern University students and faculty team members who participated in the E-ROBOT competition and helped develop and test PARIS.

\bibliographystyle{IEEEtran}

\bibliography{IEEEabrv, library}

\addtolength{\textheight}{-12cm}

\end{document}